\documentclass[10pt,twocolumn,letterpaper]{article}
\pdfoutput=1 
\usepackage{btas}
\usepackage{times}
\usepackage{epsfig}
\usepackage{graphicx}
\usepackage{amsmath}
\usepackage{amssymb}
\usepackage{pbox}
\usepackage{url}
\usepackage{rotating}
\usepackage{subcaption}

\newcommand{\norm}[1]{\lVert #1 \rVert}
\newcommand{\abs}[1]{\lvert #1 \rvert}
\newcommand{\inner}[2]{\langle #1 , #2 \rangle}
\newcommand{\minimize}{\text{minimize}}
\newcommand{\st}{\text{subject to}}
\newcommand{\R}{\mathbb{R}}
\newcommand{\clusterwidth}{0.15\linewidth}
\newcommand{\MACNN}{CNNAA}

\DeclareMathOperator{\diag}{diag}

\btasfinalcopy 

\ifbtasfinal\fi
\begin{document}

\title{Convolutional Neural Networks for Attribute-based Active Authentication On Mobile Devices}

\author{Pouya Samangouei\\
University of Maryland, College Park\\
MD, USA\\
{\tt\small pouya@umiacs.umd.edu}
\and
Rama Chellappa\\
University of Maryland, College Park\\
MD, USA\\
{\tt\small rama@umiacs.umd.edu}
}

\maketitle
\thispagestyle{empty}

\begin{abstract}
We present a Deep Convolutional Neural Network (DCNN) architecture for the task of continuous authentication on mobile devices. To deal with the limited resources of these devices, we reduce the complexity of the networks by learning intermediate features such as gender and hair color instead of identities. We present a multi-task, part-based DCNN architecture for attribute detection that performs better than the state-of-the-art methods in terms of accuracy. As a byproduct of the proposed architecture, we are able to explore the embedding space of the attributes extracted from different facial parts, such as mouth and eyes, to discover new attributes. Furthermore, through extensive experimentation, we show that the attribute features extracted by our method outperform the previously presented attribute-based method and a baseline LBP method for the task of active authentication. Lastly, we demonstrate the effectiveness of the proposed architecture in terms of speed and power consumption by deploying it on an actual mobile device.

\end{abstract}

\section{Introduction}
Mobile devices, such as cellphones, tablets, and smart watches have become  inseparable parts of people's lives. The users often store important information such as bank account details or credentials to access their sensitive accounts on their mobile phones. According to the survey in \cite{nq}, nearly half of the users do not use any form of authentication mechanism for their phones because of the frustrations made by these methods. Even if they do, the initial password-based authentication can be compromised and thus it cannot continuously protect the personal information of the users. \begin{figure}[ht]
\centering
\includegraphics[width=\linewidth]{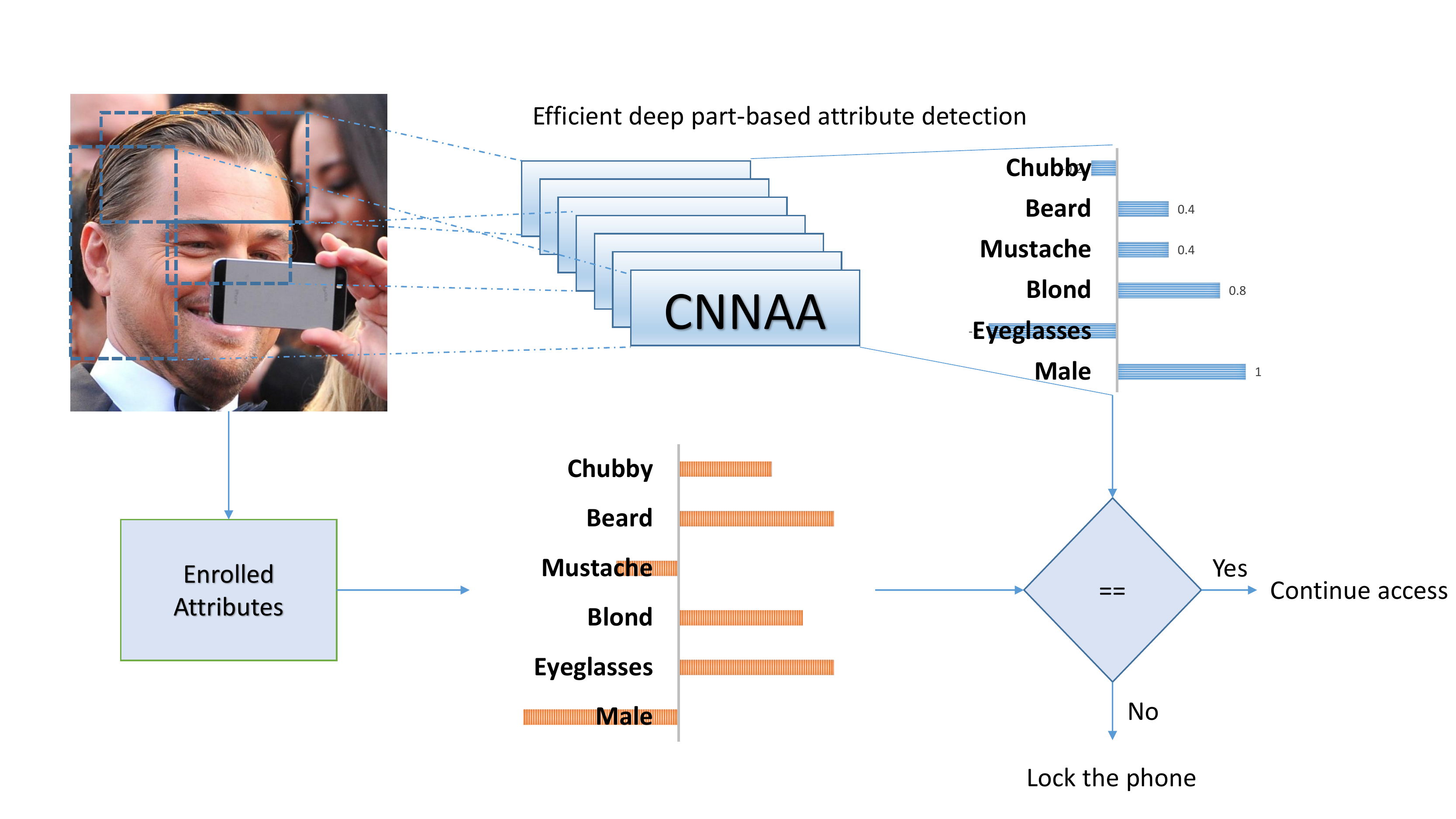}
\caption{The overview of our method. To authenticate the users, we extract facial attributes by extracting the face parts and feeding them to Convoulutional Neural Network for facial Attribute-based Acitve authentication \MACNN, which is an ensemble of efficient multi-task DCNNs.}
\label{fig:overview}
\end{figure}

To mitigate this issue and make mobile devices more secure, different active authentication (AA) methods have been proposed over the past five years to continuously authenticate the user after he/she unlocks the device. \cite{Touchalytics}, \cite{Continuous_HST2012}, \cite{Heng_WACV2015}, and \cite{antal2016biometric} proposed to continuously authenticate the users based on their touch gestures or swipes. Gait as well as device movement patterns measured by the smartphone accelerometer were used in \cite{MobileGait}, \cite{zhong2014sensor}, \cite{context_AA} for continuous authentication. Stylometry, GPS location, web browsing behavior, and application usage patterns were used in \cite{Lex_stylometry} for active authentication.
Face-based continuous user authentication on mobile devices has also been proposed in \cite{face_eye_mobile}, \cite{UMDAA}, \cite{mobio}, and \cite{mybtas,faaa}. Different modalities such as speech \cite{mobio}, Gait \cite{Jain_AA_ICB2015}, touch \cite{Heng_FG2015_Fusion} have been fused with faces.

State-of-the-art methods for face recognition employ Deep Convolutional Neural Networks (DCNN) \cite{deepface}, \cite{vgg-face}, \cite{facenet}, \cite{deepid}. The previously proposed architectures have many parameters to account for the huge variabilities of facial identities in different conditions. Thus they are not efficient to be used on mobile devices. For instance, it has been shown in \cite{sayantan} that DCNN with an architecture similar to AlexNet \cite{alexnet} can drain the battery very fast.

Facial attributes are also referred to as ``soft biometrics'' in the literature \cite{softbiometric}, and a few of them were used in \cite{niinuma2010soft} to boost the continuous authentication performance. The recent method of Attribute-based Continuous Authentication (ACA) \cite{mybtas,faaa} shows that large number of attributes can give good authentication results on mobile phones on their own. They are semantic features which are easier to learn than facial identities. Also, if they are learned on facial parts, they become less complex. By leveraging these two qualities, we design efficient CNN architectures suitable for mobile devices. As for many other tasks, CNNs are proved to be very accurate for attribute detection \cite{celeba},  \cite{zhang2014panda}, \cite{hassner}. The overview of our method is presented in Figure \ref{fig:overview}.

The main contribution of this paper is the design of feasible deep architectures for continuous authentication on mobile devices. We also obtain state-of-the-art results from the proposed multi-task, part-based deep architecture for the task of facial attribute extraction. We show that unsupervised subspace clustering of the shared embeddings gathers the semantically related attributes in the same cluster. In addition, we show improvements in attribute-based AA on two publicly available mobile datasets using the attributes from our approach and also the ``discovered'' attributes by clustering. Finally, we demonstrate the feasibility of the proposed method for mobile devices by testing the speed and power usage calculations on a commercial mobile device.

The rest of this paper is organized as follows. We present the attribute detection models in Section \ref{sec:attributes}. In Section \ref{sec:activeauth}, we show that the learned attribute models are effective for active authentication. In Section \ref{sec:performance}, we evaluate the performance of the proposed architecture using a generic Android device.

\section{Attributes}
\label{sec:attributes}
In the mobile setting, there is a trade-off between hardware constraints such as battery life, and accuracy of the models. We design our models with the goal of balancing this trade-off. To do so, we move from a computationally expensive but specialized models to a computationally cheaper but adequately accurate model. 

\begin{table}[ht]
 \centering 
 \resizebox{\linewidth}{!}{
 \begin{tabular}{|c|c|c|c|c||c|c|c|} 
 \hline 
Attribute &\begin{turn}{90} DeepMulti-\MACNN \end{turn}& \begin{turn}{90}WideMulti-\MACNN\end{turn}&\begin{turn}{90} DeepBinary-\MACNN\end{turn}& \begin{turn}{90}WideBinary-\MACNN\end{turn}& \begin{turn}{90}FaceTracer\end{turn}&\begin{turn}{90} PANDA\end{turn}&\begin{turn}{90} LNet+ANet\end{turn} \\ 
  \hline 
5 o Clock Shadow & \textbf{93}& 93& 91& 89& 85& 88& 91 \\ 
 \hline 
Arched Eyebrows & 81& 82& 82& \textbf{83}& 76& 78& 79 \\ 
 \hline 
Attractive & 81& 81& 81& \textbf{82}& 78& 81& 81 \\ 
 \hline 
Bags Under Eyes & 83& \textbf{84}& 83& 82& 76& 79& 79 \\ 
 \hline 
Bald & \textbf{99}& 99& 96& 98& 89& 96& 98 \\ 
 \hline 
Bangs & \textbf{95}& 95& 94& 94& 88& 92& 95 \\ 
 \hline 
Big Lips & 67& \textbf{70}& 69& 67& 64& 67& 68 \\ 
 \hline 
Big Nose & 82& \textbf{83}& 78& 78& 74& 75& 78 \\ 
 \hline 
Black Hair & 86& 86& \textbf{88}& 87& 70& 85& 88 \\ 
 \hline 
Blond Hair & \textbf{95}& 95& 94& 94& 80& 93& 95 \\ 
 \hline 
Blurry & \textbf{95}& 95& 92& 80& 81& 86& 84 \\ 
 \hline 
Brown Hair & \textbf{86}& 86& 86& 84& 60& 77& 80 \\ 
 \hline 
Bushy Eyebrows & \textbf{92}& 92& 89& 89& 80& 86& 90 \\ 
 \hline 
Chubby & \textbf{95}& 95& 87& 91& 86& 86& 91 \\ 
 \hline 
Double Chin & \textbf{96}& 96& 89& 93& 88& 88& 92 \\ 
 \hline 
Eyeglasses & \textbf{99}& 99& 99& 99& 98& 98& 99 \\ 
 \hline 
Goatee & \textbf{97}& 97& 93& 96& 93& 93& 95 \\ 
 \hline 
Gray Hair & \textbf{98}& 98& 92& 97& 90& 94& 97 \\ 
 \hline 
Heavy Makeup & 90& 90& 90& \textbf{91}& 85& 90& 90 \\ 
 \hline 
High Cheekbones & 86& 85& \textbf{87}& 87& 84& 86& 87 \\ 
 \hline 
Male & \textbf{98}& 97& 97& 98& 91& 97& 98 \\ 
 \hline 
Mouth Slightly Open & 93& 93& \textbf{94}& 94& 87& 78& 92 \\ 
 \hline 
Mustache & \textbf{97}& 96& 88& 95& 91& 87& 95 \\ 
 \hline 
Narrow Eyes & \textbf{87}& 87& 83& 81& 82& 73& 81 \\ 
 \hline 
No Beard & 95& 95& 95& \textbf{96}& 90& 75& 95 \\ 
 \hline 
Oval Face & 72& \textbf{73}& 73& 70& 64& 72& 66 \\ 
 \hline 
Pale Skin & \textbf{97}& 97& 93& 94& 83& 84& 91 \\ 
 \hline 
Pointy Nose & 75& 73& 75& 74& 68& \textbf{76}& 72 \\ 
 \hline 
Receding Hairline & \textbf{92}& 92& 88& 90& 76& 84& 89 \\ 
 \hline 
Rosy Cheeks & \textbf{94}& 94& 87& 91& 84& 73& 90 \\ 
 \hline 
Sideburns & 95& 95& 95& \textbf{96}& 94& 76& 96 \\ 
 \hline 
Smiling & \textbf{92}& 92& 92& 92& 89& 89& 92 \\ 
 \hline 
Straight Hair & \textbf{79}& 79& 78& 79& 63& 73& 73 \\ 
 \hline 
Wavy Hair & 71& 73& \textbf{82}& 81& 73& 75& 80 \\ 
 \hline 
Wearing Earrings & 83& 84& 86& 79& 73& \textbf{92}& 82 \\ 
 \hline 
Wearing Hat & 98& 98& 98& 98& 89& 82& \textbf{99} \\ 
 \hline 
Wearing Lipstick & 92& 92& \textbf{93}& 93& 89& 93& 93 \\ 
 \hline 
Wearing Necklace & \textbf{86}& 86& 71& 71& 68& 86& 71 \\ 
 \hline 
Wearing Necktie & 95& \textbf{96}& 93& 95& 86& 79& 93 \\ 
 \hline 
Young & 87& 87& 87& \textbf{88}& 80& 82& 87 \\ 
 \hline \hline
Average & \underline{89.4}& \textbf{89.5}& 87.7& 87.9& 81.1& 83.6& 87.3 \\ 
 \hline 
\end{tabular}
}
\caption{The accuracy comparison of attribute detection methods. Our multi-task part-based architectures perform better than previously proposed methods and also single-task networks.}
\label{tab:attr_acc}
\end{table}

We train and test four different sets of DCNNs, in total 100 of them, for the task of attribute classification on a set of face regions. We crop the functional face regions using landmarks detected by \cite{asthana}. These face regions can be seen in Table \ref{tab:parts}. For each part, first we find the maximum size of the window in the training dataset, then we crop the regions by putting the center of the face part at the center of the cropped window to avoid any scaling.

\subsection{Network architecture}
The two proposed architectures of the Deep Convolutional Neural Network for facial Attribute-based Active authentication (Deep-\MACNN) and Wide-\MACNN\ can be found in Table \ref{tab:networks}. The four sets of models compared are: BinaryDeep-\MACNN\ and BinaryWide-\MACNN, which are single task networks, MultiDeep-\MACNN\ and MultiWide-\MACNN, which are multi-task networks. First we describe the shared configuration that is used to train these networks and then the ones that are specific to each type.

\begin{table}
\begin{tabular}{|c|c||c|c|}
\hline
\multicolumn{2}{|c|}{*Wide-\MACNN} & \multicolumn{2}{|c|}{*Deep-\MACNN} \\
\hline
input & $w \times h \times 3$ & input & $w \times h \times 3$ \\
\hline
type & patch size & type & patch size\\
\hline
conv\_relu & $7\times7\times128$ & conv\_relu & $7\times7\times32$ \\
\hline
\multicolumn{4}{|c|}{maxpool $3\times3/2$(stride)} \\
\hline
conv\_relu & $5\times5\times128$ & conv\_relu & $5\times5\times32$ \\
&& conv\_relu & $5\times5\times32$ \\
&& conv\_relu & $5\times5\times32$ \\
\hline
\multicolumn{4}{|c|}{maxpool $3\times3/2$} \\
\hline
conv\_relu & $3\times3\times128$ & conv\_relu & $3\times3\times32$ \\
&& conv\_relu & $3\times3\times32$ \\
&& conv\_relu & $3\times3\times32$ \\
&& conv\_relu & $3\times3\times32$ \\
\hline
\multicolumn{4}{|c|}{maxpool $3\times3/2$} \\
\hline
FC\_relu & $dim \times 128$ & FC\_relu & $dim \times 64 $\\
FC\_relu & $128 \times 128$ & FC\_relu & $64 \times 32 $\\
\hline
\multicolumn{4}{|c|}{logits $Num\_Attr\times2$} \\
\hline
\multicolumn{4}{|c|}{Softmax loss} \\
\hline
\end{tabular}
\caption{The architectures of our networks. The number of parameters depends on the face region that they operate on and can be found in Table \ref{tab:performance}.}
\label{tab:networks}
\end{table}

\newcommand{\partwidth}{0.4}
\newcommand{\partwidthh}{0.3}

\newcommand{\ac}{c}
\begin{table*}[t]
\centering

\begin{tabular}{\ac \ac \ac \ac \ac \ac}

face part
& \includegraphics[scale=\partwidth]{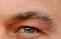} 
&\includegraphics[scale=\partwidth]{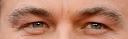} 
& \includegraphics[scale=\partwidth]{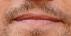} 
& \includegraphics[scale=\partwidth]{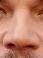} 
&  \includegraphics[scale=0.3]{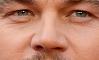} \\
\hline
No. of Attributes & 10 &10 &10 &7 &16 \\
\hline
Size & $53\times39$ & $115\times41$ & $65\times38$ & $40\times56$ & $90\times62$\\

face part& \includegraphics[scale=\partwidth]{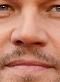}
& \includegraphics[scale=\partwidthh]{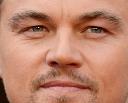}
& \includegraphics[scale=\partwidthh]{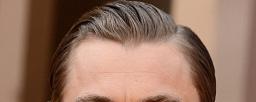}
& \includegraphics[scale=\partwidthh]{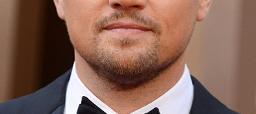}
& \includegraphics[scale=\partwidthh]{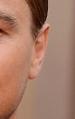}\\
\hline
No. of Attributes &15 &21 &15 &15 &14 \\
\hline
Size & $55\times82$ & $115\times107$ & $128\times52$ & $128\times 45$ & $62\times100 $\\
\hline

\end{tabular}
\caption{The face regions that are extracted by cropping around the landmark points and their corresponding number of attributes. A Multi*-\MACNN\ that operates on a face crop has ``No. of Attributes'' tasks.}
\label{tab:parts}
\end{table*}

\paragraph{Shared configuration} All of these 100 networks, 20 Multi*-\MACNN, 2 networks per 10 parts, and 80 Binary*-\MACNN, 2 networks per attribute on full face, are trained on the publicly available CelebA \cite{celeba} dataset. It has 200 thousands images of 10 thousands identities, each with 40 attribute labels. It is divided into 160k training, 20k development, and 20k test images.  The DCNNs are trained using the recently released Tensorflow \cite{tensorflow} which also has a mobile implementation. All of the networks are initialized with random weights and are trained with the same policy. The Adam optimizer is used to train all of these networks since it incorporates the adaptive learning rate update step, and performs well without a careful fine tuning of the learning parameters \cite{adam}. Subsequent fine tuning can give better results. Early stopping \cite{early_stopping} using the accuracy on the development set is used to select the final model for each network. The inputs are colored images that are randomly flipped and also their contrast and Gamma are randomly changed to augment the data to prevent over-fitting.

Due to the nature of the attributes, most of them have an unequal number of positive and negative labels. Extra care has been taken to make sure the networks are not biased toward one class with the help of data augmentation and stochastic optimization.

\paragraph{Binary*-\MACNN} The binary networks are for a single task and are trained by the labels of one single attribute. The input face images are aligned to a canonical coordinate. To balance the training data, the class with the lower number of training data is distorted and added to the input queue so that the number of images for each class is equal. Then the data is shuffled and fed in batches to the training algorithm. The softmax cross entropy loss $l_B$ in (\ref{eq:binary}) is used to train these binary networks
\begin{equation} 
\begin{aligned} 
l_B(w) = & \frac{1}{N}\sum_{i=1}^N (1-y_i) \log p(y_i=0|w) \\
 & + y_i \log p(y_i=1|w)
\end{aligned}
\label{eq:binary}
\end{equation}
where $N$ is the batch size, $y_j \in \{0,1\}$ is the attribute presence label, $p(y=j|w) = \frac{\exp{(f_j^w(x))}}{\sum_{i=0}^1\exp{(f_i^w(x))}}$ where $f_i^w(x)$ is the logits of the $i$th output neuron of the network with weights $w$.

\paragraph{Multi*-\MACNN} The Multi* networks have the same complexity as binary models but predict multiple attributes at once. The face parts and the number of attributes that are assigned to them can be found in Table \ref{tab:parts}. For each part, the corresponding network has an output layer that contains neurons for each attribute that is assigned to that face part. We use the softmax cross entropy loss for part $q$ as specified below:
\begin{equation}
\begin{aligned}
l^{q}(w) = & \frac{1}{N^q}\sum_{a=1}^{N^{q}} \frac{1}{n_a^q}\sum_{i=1}^{n_a^{q}} (1-y^a_i) \log p(y_i=0|w) \\
 & + y_i^a \log p(y^a_i=1|w)
\end{aligned}
\end{equation}
where $N^q$ is the number of attributes assigned to part $q$. $n_a^{q}$ is the number of images with the $a$th attribute of part $q$ in the current batch. $y_i^a \in \{0,1\}$ is 1 if the $i$th image has the $a$th attribute and N is the batch size. $p(y^a_i=1|w)$ is the same as Eq \ref{eq:binary}.

To deal with the class ratio imbalance of the attributes, we shuffle the training data in a way that the network sees the rare class for each attribute frequently.  For example, for the attribute ``Mustache'', the positive class is the rare one since most of the $202k$ images do not have this attribute. To handle this imbalance, a queue is created for each attribute and images that have the rare class are added to that queue. A queue is also created for images with all the attributes belonging to the major class. Then all of the queues are shuffled. We treat each queue as a circular buffer so that the training batches are created by sampling with replacement from one of these queues at random.  Also, each time the images are distorted differently.

After training all of the networks separately, we train a single linear Support Vector Machines (SVMs) classifier per attribute over the embeddings of these networks. For each attribute, we only take the embeddings of the relevant parts. For instance, for the attribute "Mustache" in the MultiDeep-\MACNN, the 32 dimensional embedding of the parts: mouth, mouth-and-nose, and mouth-and-chin are taken and concatenated together. The SVMs are trained using the training set of CelebA \cite{celeba} and fine tuned on its development set. 

\subsection{Comparison of attribute detection methods}

\begin{figure*}[ht]
\centering
\begin{tabular}{c |c c c c c}
\begin{turn}{90}Forehead and hair\end{turn}&
\includegraphics[width=\clusterwidth]{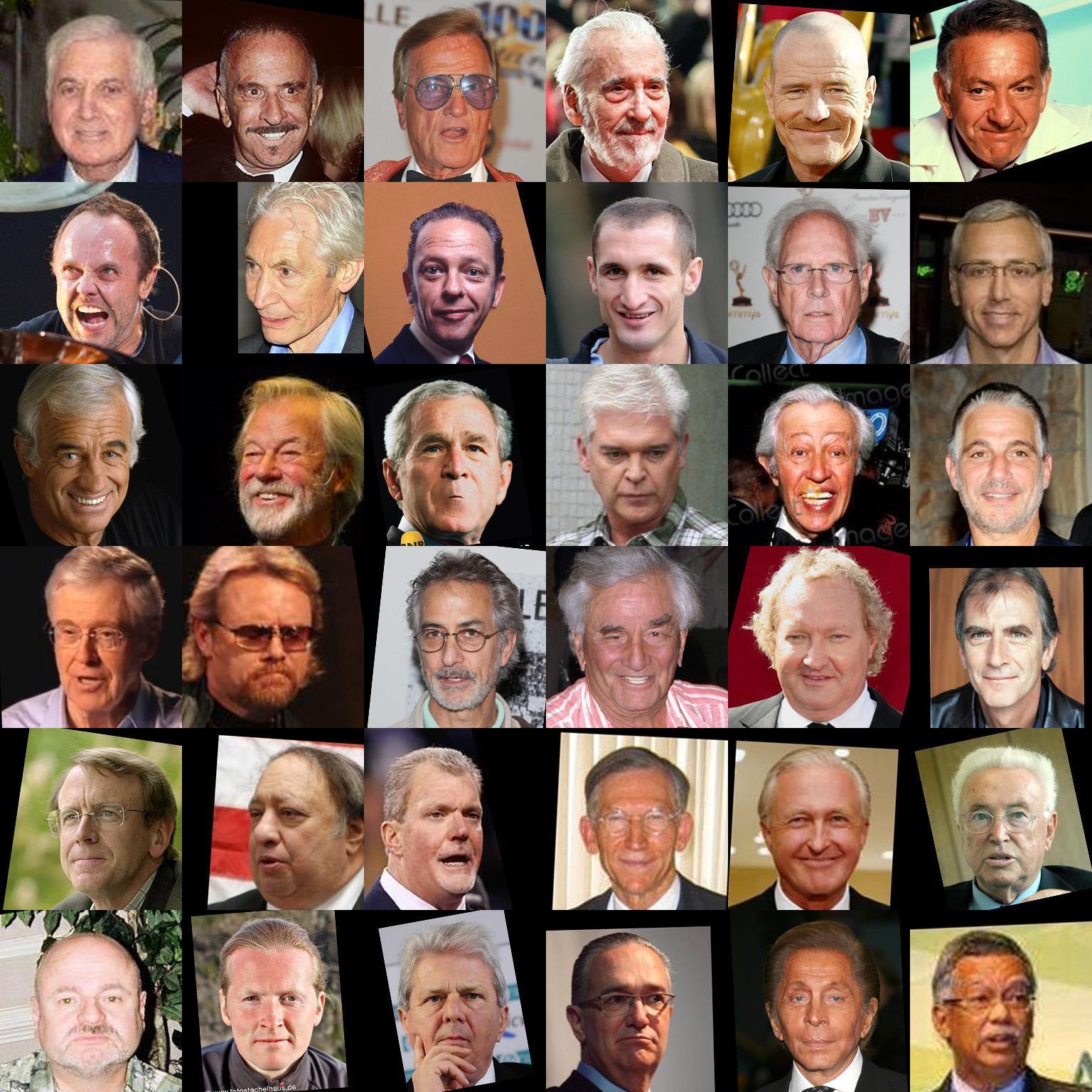} & \includegraphics[width=\clusterwidth]{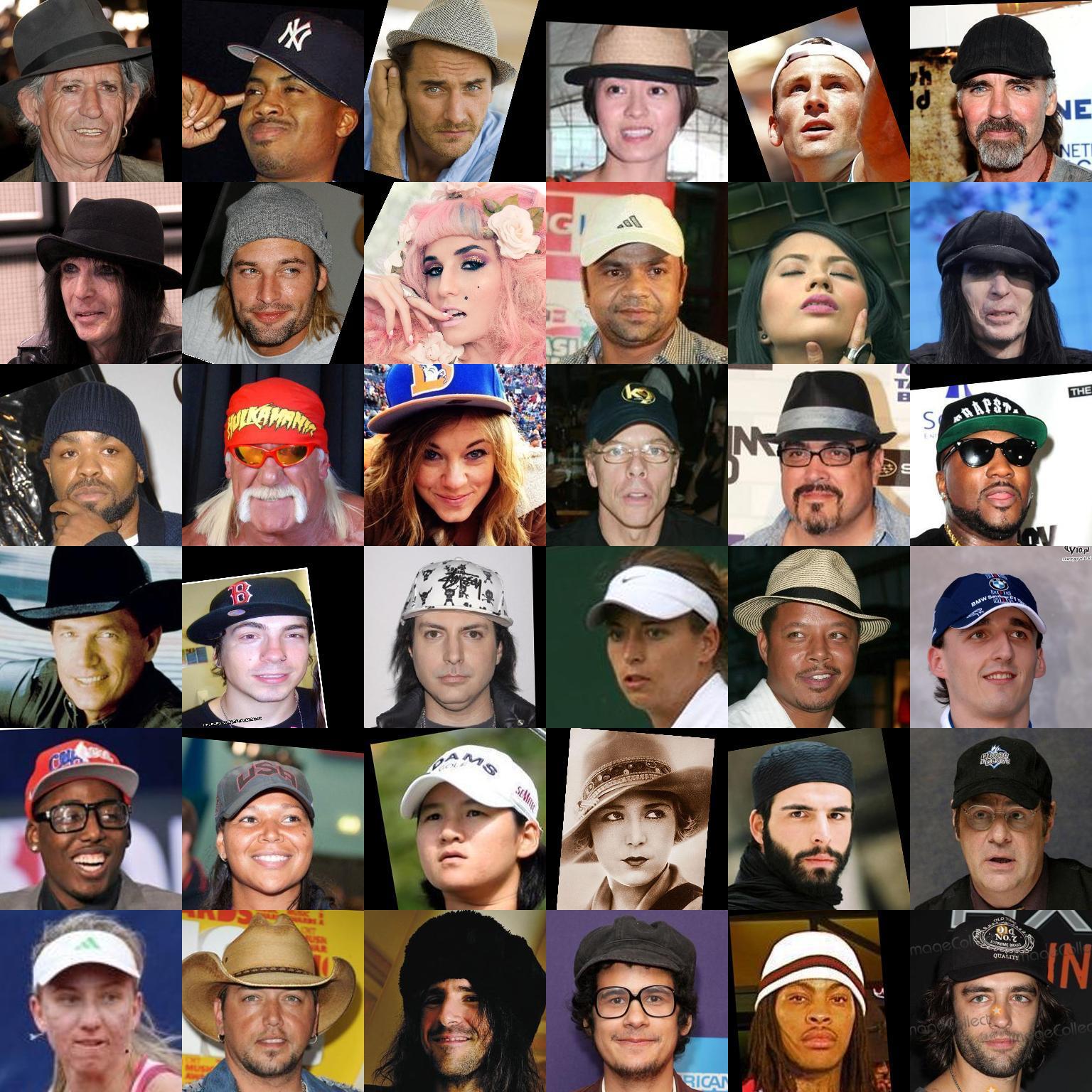} & \includegraphics[width=\clusterwidth]{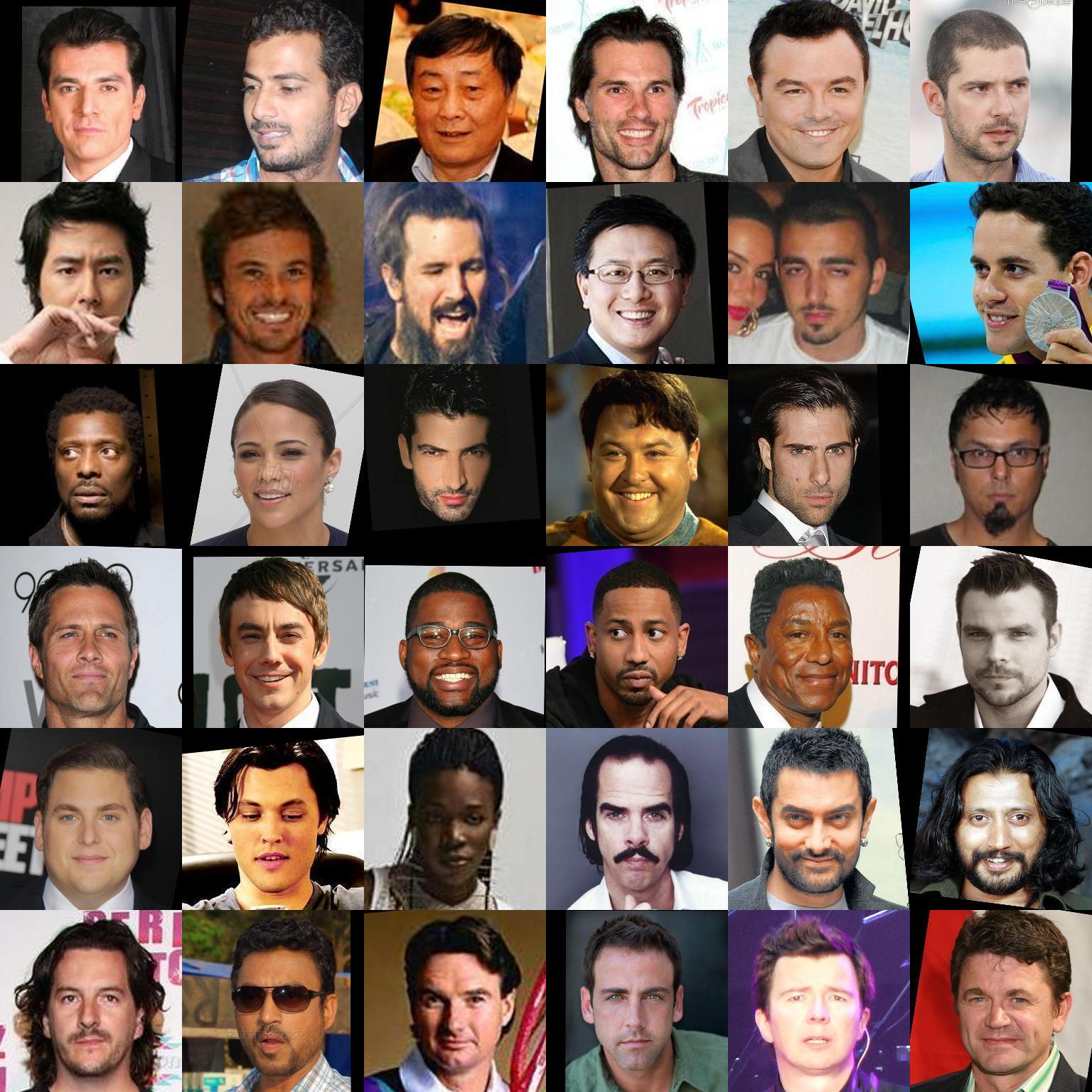} & \includegraphics[width=\clusterwidth]{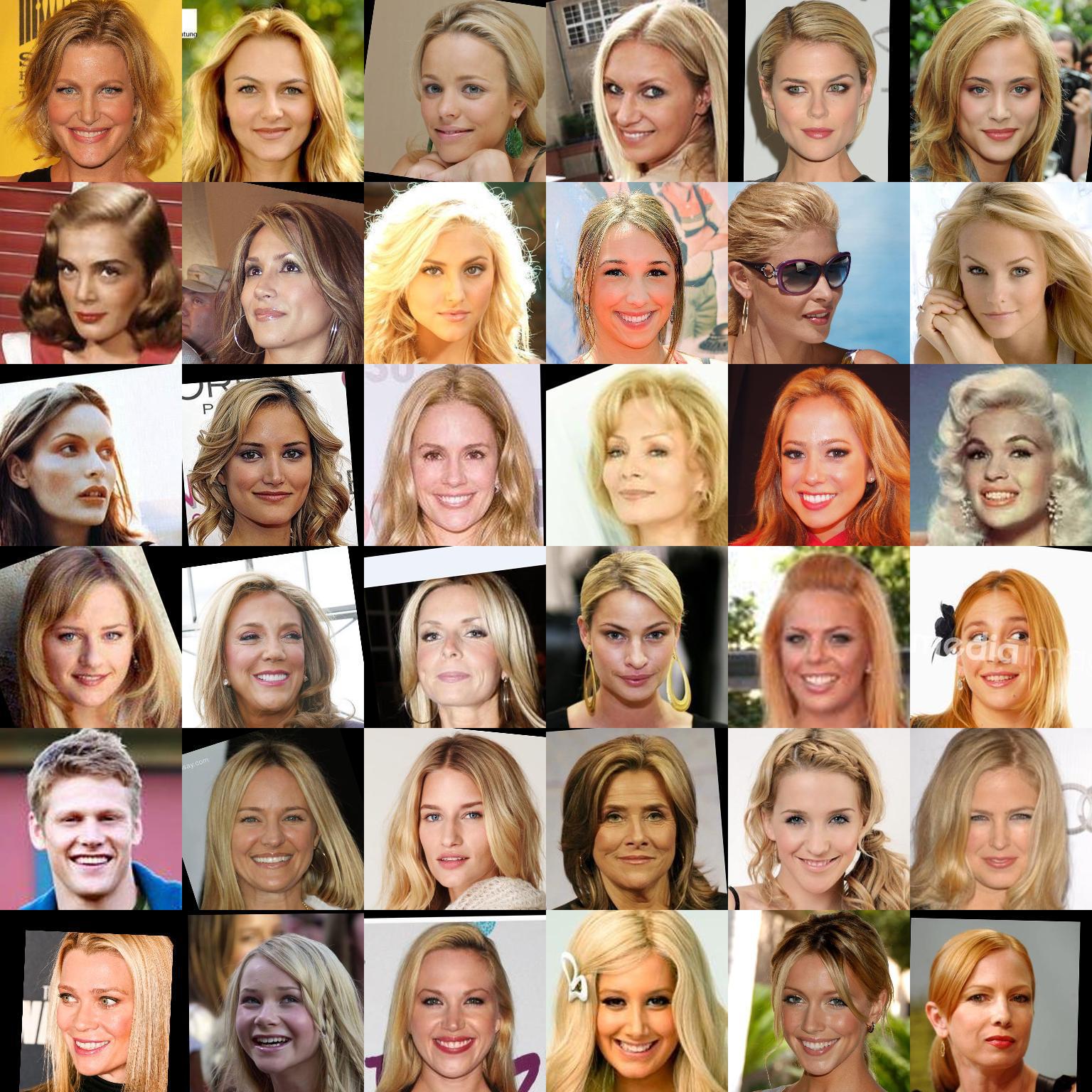} & \includegraphics[width=\clusterwidth]{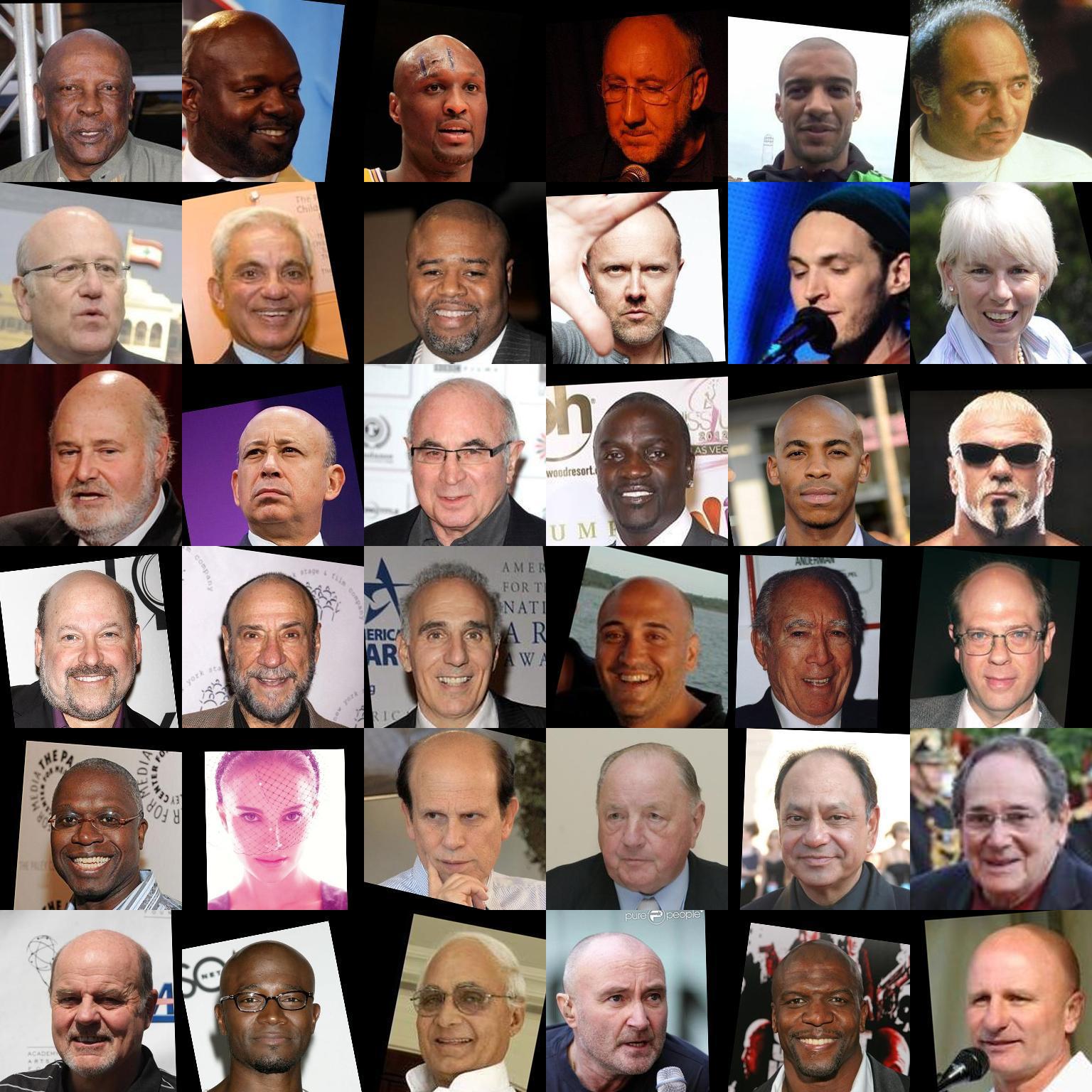} \\
&(a) & (b) & (c) & (d) & (e) \\ 
\begin{turn}{90}Mouth and Chin\end{turn} &
\includegraphics[width=\clusterwidth]{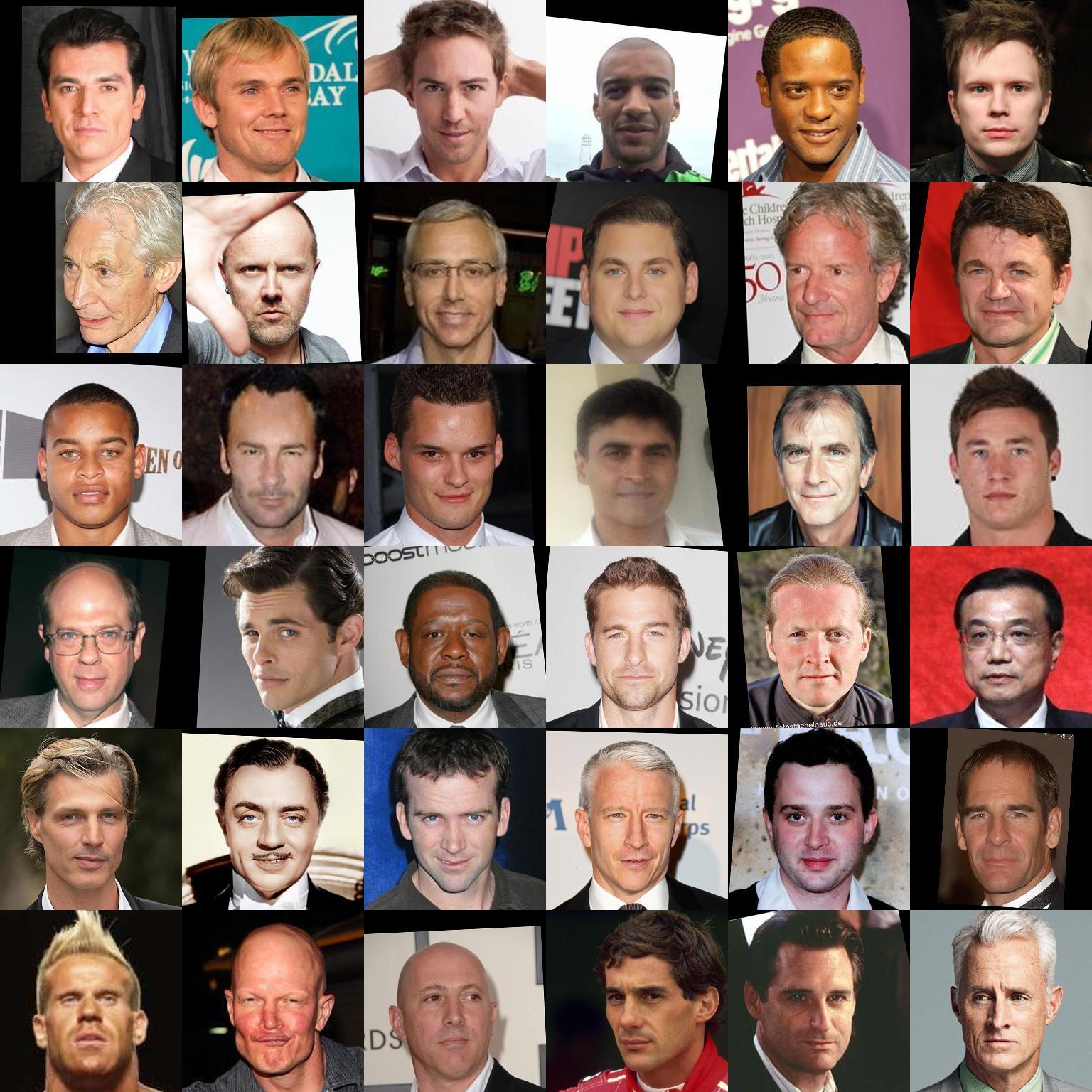} & \includegraphics[width=\clusterwidth]{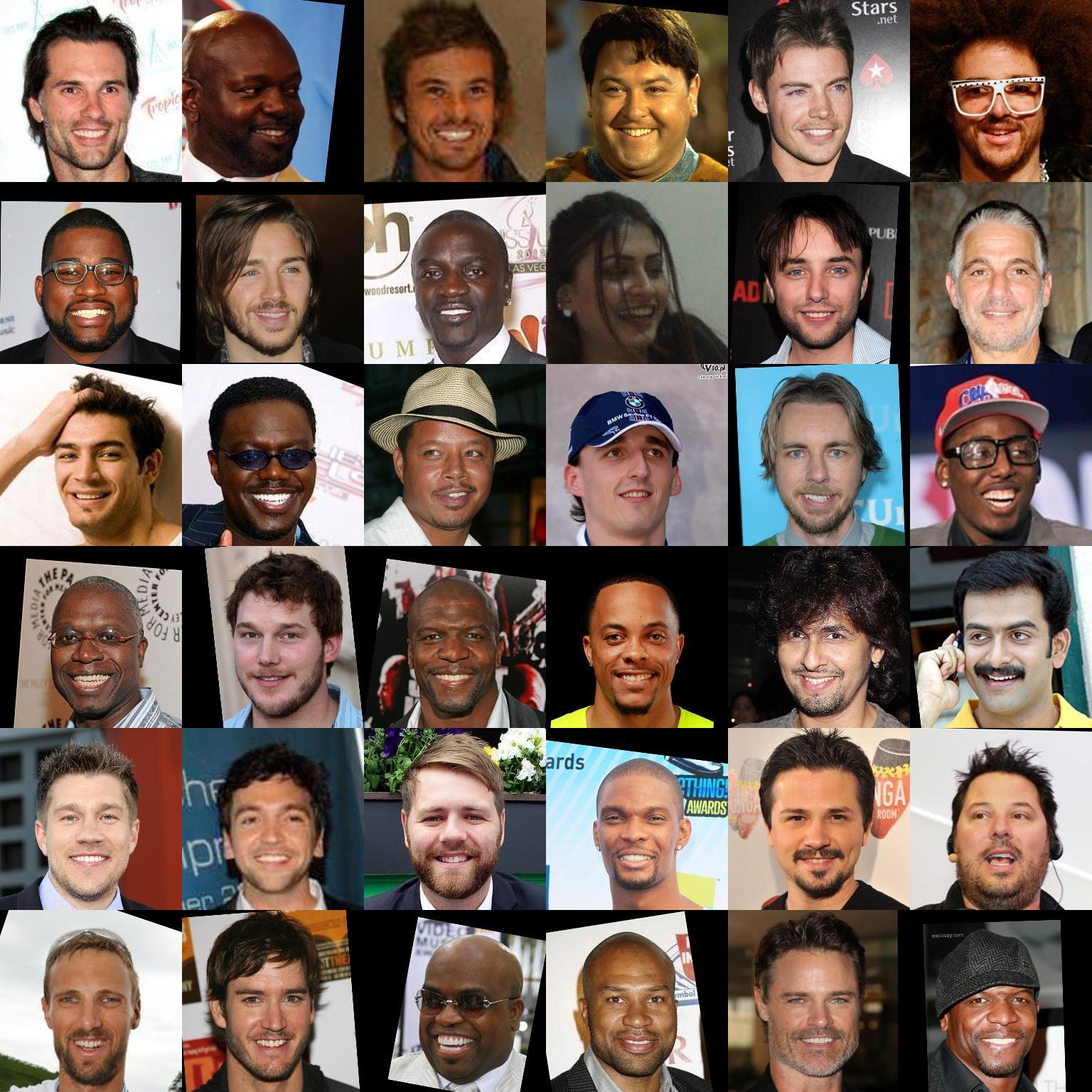} & \includegraphics[width=\clusterwidth]{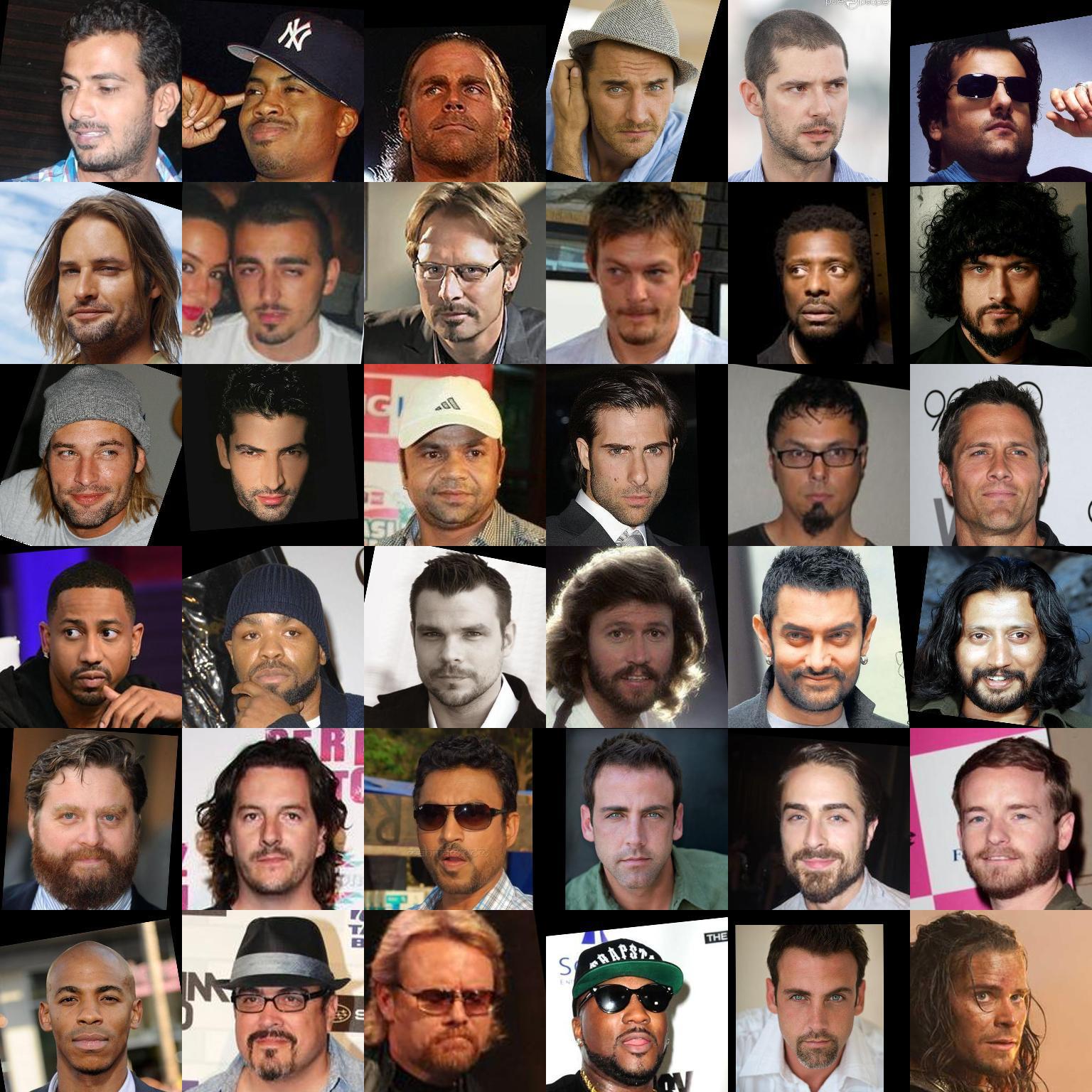} & \includegraphics[width=\clusterwidth]{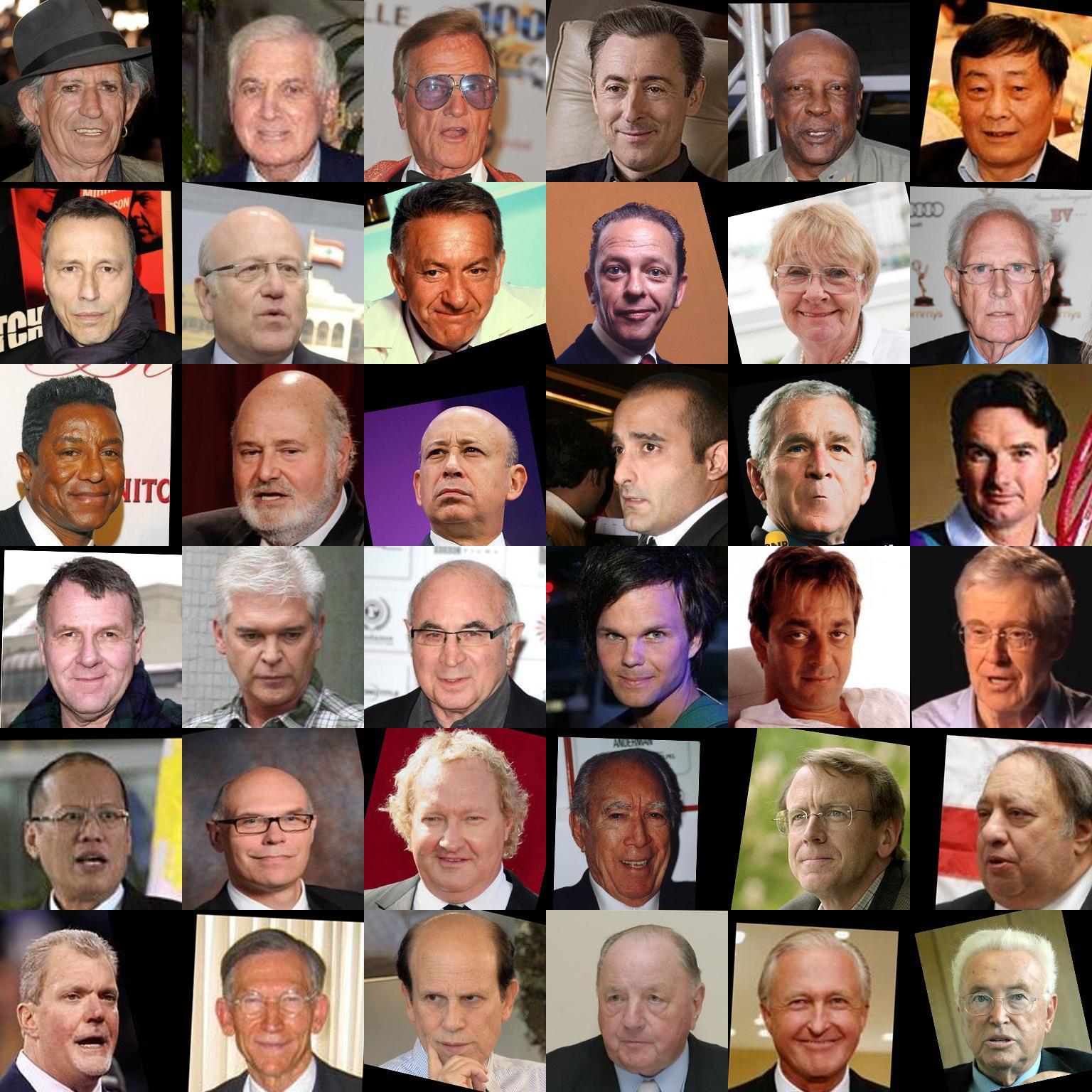} & \includegraphics[width=\clusterwidth]{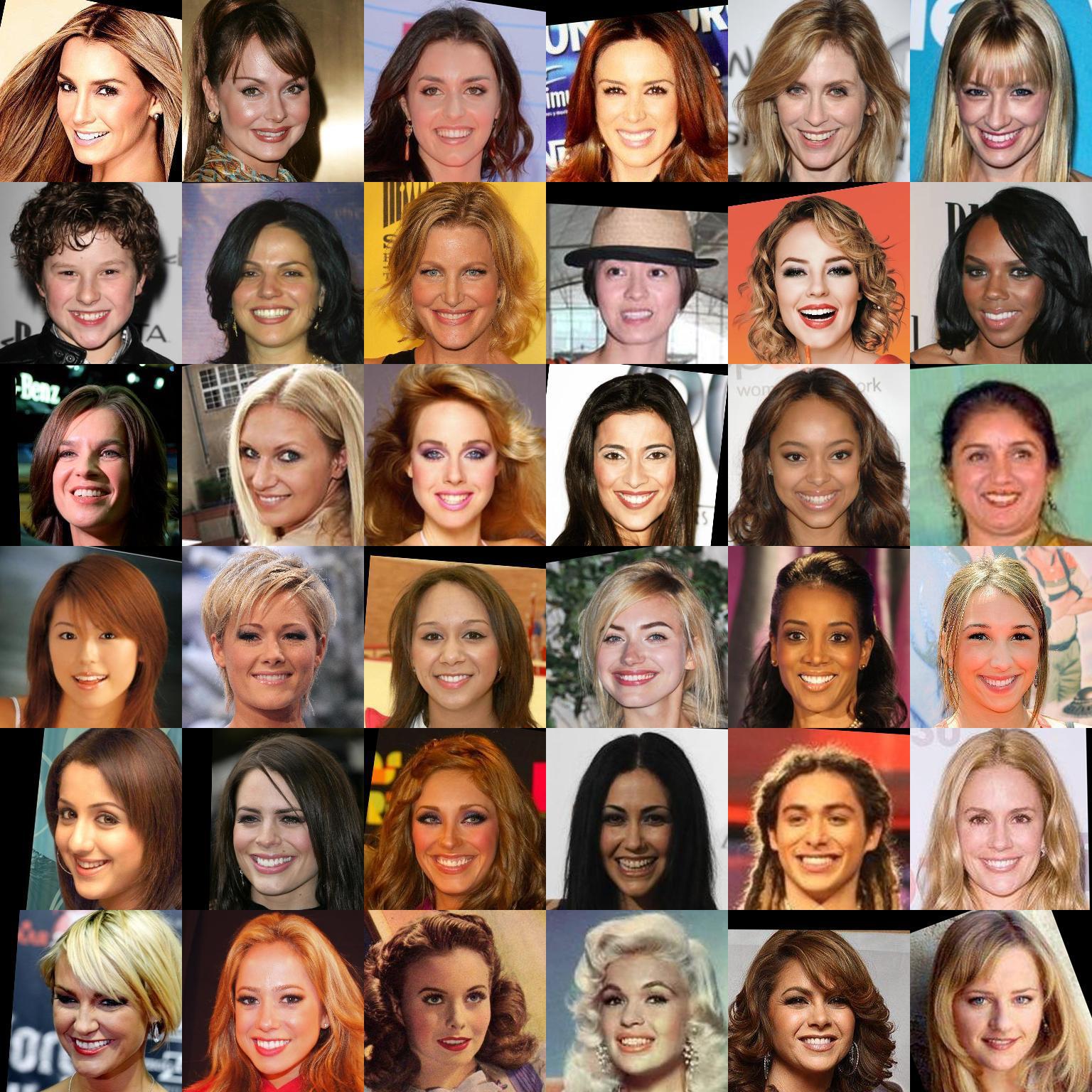} \\
& (a) & (b) & (c) & (d) & (e) 
\end{tabular}
\caption{Sample images from subspace clustering of face part embedding in attribute space. Zoom in to see the clusters. }
\label{fig:clustering}
\end{figure*}

We compare our proposed networks with FaceTracer \cite{facetracer}, PANDA \cite{zhang2014panda}, and CelebA \cite{celeba} attribute networks. These models capture a broad spectrum of possible automatic attribute detection models.

FaceTracer \cite{facetracer} attribute classifiers are trained by extracting traditional low-level features like HOG and the color histogram from aligned face parts by incrementally finding the best set of features and training the SVMs on the selected features and parts for attribute detection. The face crops are extracted from the ground truth landmarks. 

PANDA employs multiple CNNs for the face parts and concatenates the outputs of the last layer and trains SVMs for each attribute. There are two differences between our network architecture and PANDA networks. First, in PANDA, all of the attributes are associated with all of the parts. Second, in our Multi*-\MACNN\ networks, the last layer is shared between all of the attributes softmax losses, but in PANDA there are two fully connected layers after the shared fully connected layer for each one. As a result, in our network, the different attributes that are associated with one network lie in the same Euclidean space of the last fully connected layer of the network. We exploit this feature in Section \ref{sec:attr_disc}.

CelebA takes a different approach by pre-training a network with face identities of CelebFaces \cite{sun2014deep} for both face verification and identification. Then features from multiple overlapping patches of the face and train SVMs for each cropped region and each attribute. To predict an attribute, the scores of SVMs are averaged. 

We also follow \cite{hassner} and train a single task network for each attribute in Binary*-\MACNN\ on the full face to compare with the most specialized model for each attribute. Table \ref{tab:attr_acc} shows the accuracy of each of these methods.

As it can be seen, our Multi*-\MACNN\ networks give equal or better results than the rest. The MultiWide-\MACNN\ architectures perform slightly better than the MultiDeep-\MACNN\ in attribute prediction. However, they are slower and consume more energy as shown in Section \ref{sec:performance}.

\subsection{Attribute discovery}\label{sec:attr_disc}
As mentioned in the previous section, our Multi*-\MACNN\ networks transform the input face regions to a shared Euclidean space for the attributes associated with that part. To further explore this Euclidean space, we perform Sparse Subspace Clustering (SSC) \cite{ssc} on 10000 points that are selected from the training portion of CelebA dataset. The intuition behind this clustering step is that the data points with the same set of attributes lie on the same side of the learned planes defined by the weights of the last layer of each network. Thus they can be represented as a sparse combination of the neighboring points. SSC uses this fact to find the clusters. Therefore by formulating the clustering problem as 
\begin{align}
\underset{C\in \R^{n\times n}}{\minimize} \ \ & \abs{C}_1+\norm{D-DC}^2_F \\
\st \ \ & \diag{(C)} = 0
\end{align}
where $D \in \R^{d\times n}$ is the data matrix containing $n$ points of dimension $d$ and $C \in \R^{n \times n}$ is the affinity matrix. To enforce the constraint, the authors of \cite{ssc} find the sparse code of each data point in a dictionary of all the points except the test point. To get the clusters they perform spectral clustering on $C$.

We find 10 clusters per face regions. The clusters corresponding to the ``Hair-Forehead'' region of the face and the ``eyes'' region can be seen in Figure \ref{fig:clustering}. As illustrated, the ``discovered'' attributes overlap with the labels that we had in the training time mostly, however, some attributes are divided into finer categories. For example, the ``Hair-Forehead'' region cluster (c) contains male images with short hair which was not seen in the labels. 

\section{Active Authentication}
\label{sec:activeauth}
We evaluate the performance of \MACNN\ for the task of active authentication using two publicly available datasets MOBIO \cite{mobio} and AA01 \cite{UMDAA}. These datasets contain videos of the users interacting with cell phones. We compare the authentication performance of our DCNN attribute detectors and discovered attributes using the baseline Local Binary Patterns \cite{lbp} and ACA \cite{mybtas,faaa} which is the only attribute-based approach for this task on mobile phones. We follow the same protocol as ACA to extract facial parts and video features. So, we average over the extracted attribute outputs for the video frames to get the video descriptors.

We cast the problem of continuous authentication as a face verification problem in which a pair of videos is given to determine whether they contain the same identity or not. To compare the performance of the algorithms, the receiver operating characteristic (ROC) curve is used. Many other measures of performance can be readily extracted from the ROC curve. The ROC curve plots the relationship between false acceptance rates (FARs) and true acceptance rates (TARs) and can be computed from a similarity matrix $S$ between gallery and probe videos. We also report the EER value where False Rejection Rate (FRR) and FAR are equal. EER value gives a good idea of the ROC curve shape, since the value $1-EER$ is where the line $x+y+1=0$ meets the ROC curve. Thus, the better the algorithm, the lower is its EER value.

We give each video frame to the \MACNN\ networks and predict the attributes with linear SVMs. For the learned attributes, we put the probabilistic output of the SVMs which are trained by LIBSVM \cite{libsvm} as our final attribute feature.  Since the attribute outputs of our models are probability values we get the similarity value $s_{i,j} = \inner{e_i}{t_j}$, where $e_i$ is the feature vector for the enrollment video and $t_i$ is the test video features.

\begin{figure}
\centering
\includegraphics[width=0.6\linewidth]{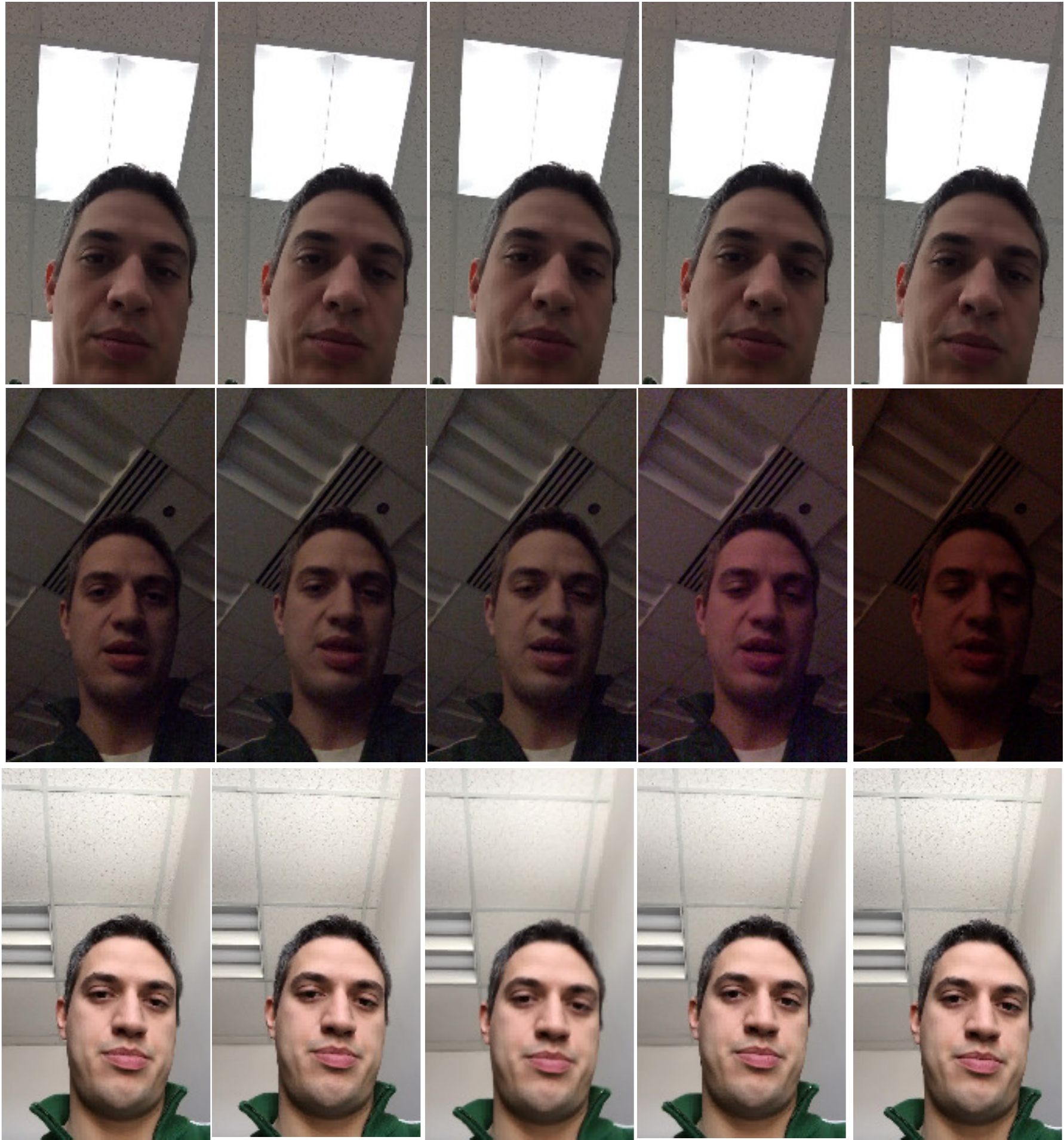}
\caption{Sample images of the three sessions of the AA01 dataset.}
\label{fig:aa_samples}
\end{figure}

To use the discovered attributes (DiscAttrs) for authentication, we extract the attribute features by a similar approach to Sparse Representation Classification \cite{src}. Each face crop from the video frame is embedded to the attribute space of MultiDeep-\MACNN. It is represented by the dictionary which we used in Section \ref{sec:attr_disc}, so that we know the cluster assignment of its atoms. We normalize all of the dictionary atoms and the embedding. Then we get each feature value by a softmax over the representation contribution of each cluster in the attribute space. To do so, we first solve
\begin{equation}
\begin{aligned}
\underset{f\in \R^{n}}{\minimize} \ & \abs{f}_1 + \norm{f-Df}_F^2
\end{aligned}
\label{eq:sparse}
\end{equation}
to get the sparse representation $f$ of the face crop of that video frame. Then we set the $i$th feature for that face crop to $p(c=i|D)$ which is calculated by
\begin{equation}
p(c=i|D) = \frac{\exp(\norm{D_{:,i}f_i})}{\sum_{k=1}^{10} \exp(\norm{D_{:,k}f_k})}
\end{equation}
where $D_{:,i}$ is the dictionary atoms of cluster $i$ and $f_i$ are the coefficients corresponding to those atoms. Thus, if $f$ is in the subspace spanned by the points in $D$ that are in cluster $i$, it will have more energy in non-zero values for those atoms. To solve (\ref{eq:sparse}) we use the Orthogonal Matching Pursuit \cite{omp} algorithm with sparsity 20. The only reason for choosing 20 is that it is less than 32 which is the embedding space dimension. This parameter could be fine tuned further to get better results. Then we concatenate all of these probability values for different face parts to create the final representation. The similarity matrix is then created in the same way as using attributes as features.
\begin{figure*}[ht]
\centering
\input{figures/aa_4}
\input{figures/mobio_roc_all} 
\input{figures/mobio_roc_mobile}
\caption{ROC curve of different experiments on AA01 \cite{UMDAA} and MOBIO \cite{mobio} dataset. (a) is the ROC curve of AA01 with all of the sessions together in gallery and probe. (b) is the ROC curve of MOBIO with all of the mobile sessions together with the last session videos as gallery and videos of the rest as probe. (c) is the ROC curve of the cross-device experiment. }
\end{figure*}
\begin{table} 
 \centering 
   \resizebox{0.8\linewidth}{!}{

 \begin{tabular}{|c|c|c|c|c|c|} 
 \hline 
 $\mathbf{e} \rightarrow \mathbf{t}$& \begin{turn}{90}ACA\end{turn}& \begin{turn}{90}LBP\end{turn}& \begin{turn}{90}MD-\MACNN\end{turn}& \begin{turn}{90}MW-\MACNN\end{turn}& \begin{turn}{90}DiscAttrs\end{turn}  \\ 
 \hline \hline
$ 1 \rightarrow 1 $ & 0.14& \underline{0.13}& \textbf{0.11}& 0.14& 0.16 \\ 
 \hline 
$ 2 \rightarrow 2$ & 0.19& 0.31& \textbf{0.18}& 0.22& \underline{0.17} \\ 
 \hline 
$3 \rightarrow 3$ & 0.16& 0.20& \textbf{0.10}& \textbf{0.10}& 0.13 \\ 
 \hline \hline
$1 \rightarrow 2,3$ & 0.38& 0.38&\textbf{0.18}& 0.25& \underline{0.23} \\ 
 \hline 
$2 \rightarrow 1,3$ & 0.31& 0.33& \textbf{0.26}& 0.30& \underline{0.31} \\ 
 \hline 
$3 \rightarrow 1,2$ & 0.31& 0.38& \textbf{0.19}& \underline{0.24}& 0.25 \\ 
 \hline \hline
Altogether & 0.30& 0.34& \textbf{0.20}& \underline{0.25}& \underline{0.25} \\ 
 \hline 
\end{tabular}
}
\caption{The EER values for the different experiments on AA01\protect\cite{UMDAA} dataset. The sessions numbers are: 1. Office light 2. Low light 3. Natural light. DiscAttrs column contains the EER values using the discovered attributes. }
\label{tab:aa_eer}
\end{table}

\subsection{Results}
To plot the ROC curves and evaluate our method, in each dataset, for each person, videos of one session are considered as the enrollment videos and the other videos as the test videos. The similarity matrix is then generated by computing the pairwise similiarity between the enrollment and the test videos. The corresponding ROC curve is plotted for each experiment.

\paragraph{AA01} AA01 is a mobile dataset with 750 videos of 50 subjects. Each subject has three sets of videos with three different lighting conditions. Each user is asked to perform a set of actions on the phone while the front camera is recording the video. The videos are captured by an iPhone 4 camera. The three lighting conditions are: office light, low light, and natural light. The sample images of this dataset in Figure \ref{fig:aa_samples} show the three different illuminations in each session. Figure \ref{fig:aa_samples} also presents some partial faces in the dataset. Each person has five videos of performing five different tasks on the phone. There is a designated enrollment video for each person. Three different experiments have been conducted on this dataset.

First, the enrollment and the test videos for all of the 50 subjects are taken from the session with the same lighting condition. The EER values of this experiment can be found in the first three rows of Table \ref{tab:aa_eer}. It can be seen that our MultiDeep-\MACNN\ has the lowest EER in all cases. This experiment reveals the discriminative power of the features when the surrounding environment is the same. 
In the second experiment, all of the enrollment videos are taken from one illumination session and the test videos from another. The EER values corresponding to this experiment are depicted in the next three rows of Table \ref{tab:aa_eer}. The performance drop in our method is $0.08$ while on average while ACA suffers $0.17$ and LBP $0.15$. The reason is that ACA attribute classifiers use low level features that are sensitive to illumination changes, but \MACNN\ is trained on a large-scale unconstrained dataset containing a lot of variations and thus gives more robust features. 

In the last experiment, all of enrollment videos of the three sessions are put in the gallery and all of the test videos in the probe to get the similarity matrix. The ROC curve corresponding to the third general experiment is plotted in Figure \ref{fig:aa_roc}. It can be seen that MultiDeep-\MACNN\ performs the best and MultiWide-\MACNN\ and the discovered attributes are tied as second best.

One explanation for the lower performance of MultiWide-\MACNN\ compared to MultiDeep-\MACNN\ is that it has many more parameters than MultiDeep-\MACNN\ according to Table \ref{tab:performance} and has overfitted to the celebrity face images distribution.

\begin{figure}[t]
\centering
\includegraphics[width=0.8\linewidth]{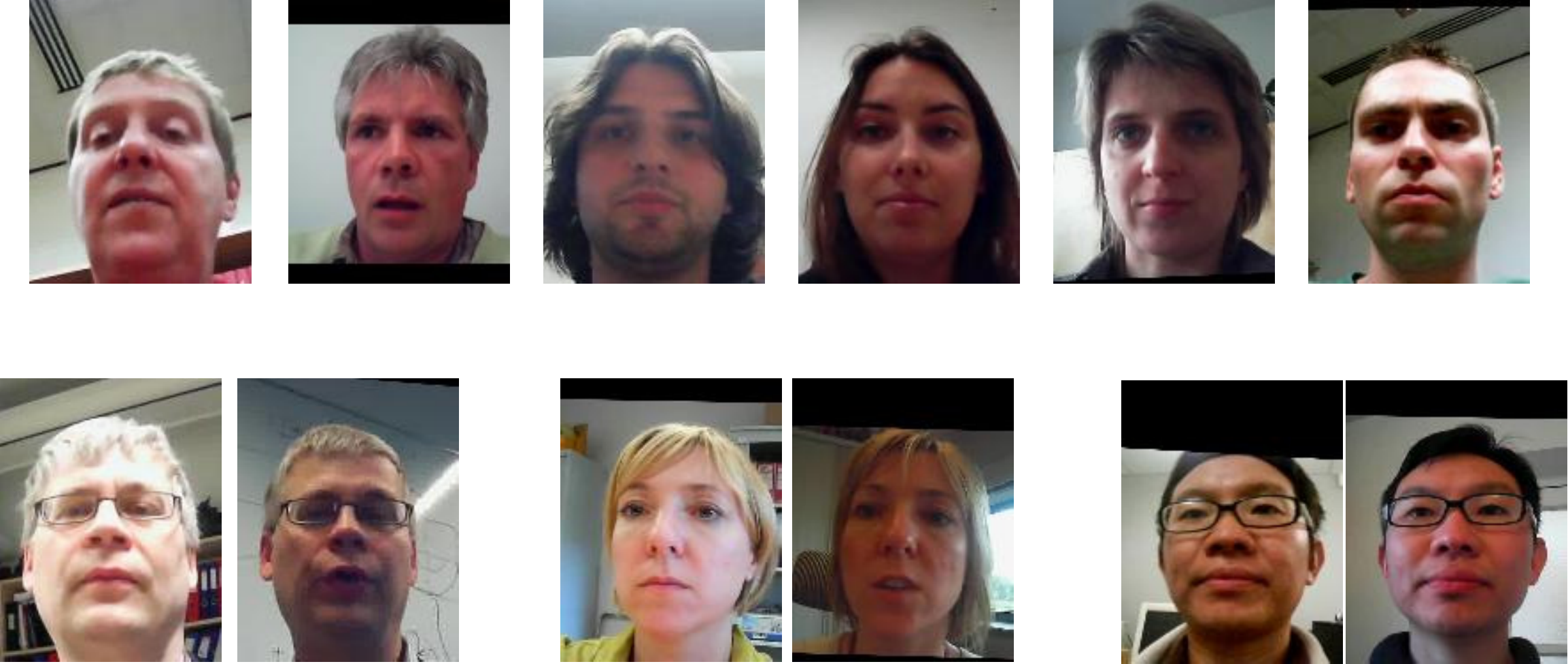}
\caption{Sample images of three sessions of the MOBIO dataset. First row images are from different sites, second row has the pairs with the same identities in two different sessions. }
\end{figure}

\paragraph{MOBIO} MOBIO \cite{mobio} is a dataset of 152 subjects. The videos are taken in six different universities across Europe. For most subjects, twelve sessions of video are captured. All of the mobile videos are captured with a Nokia N93i. The first session's videos are also recorded with a 2008 MacBook laptop. We perform two experiments on this dataset. We take the $12$th session videos as our training videos since they are the mostly available videos across the dataset.

In the first experiment, we just consider videos that are taken by the mobile device. We show the EER values for the mobile videos of the subjects within each site, as well as all of the videos together in Table \ref{tab:mobio_eer}. This experiment is similar to the ``Altogether'' experiment of AA01 dataset since the environment conditions for enrollment videos and test videos can be the same or different. The ROC curve for this experiment is plotted in Figure \ref{fig:mobio_roc_all}.

\begin{table} 
 \centering 
 \resizebox{0.8\linewidth}{!}{
 \begin{tabular}{|c|c|c||c|c|c|} 
 \hline 
 & \begin{turn}{90}ACA\end{turn}& \begin{turn}{90}LBP\end{turn}& \begin{turn}{90}MD-\MACNN\end{turn}& \begin{turn}{90}MW-\MACNN\end{turn}& \begin{turn}{90}DiscAttrs\end{turn} \\ 
 \hline 
but & 0.26& 0.36& \textbf{0.19}& \underline{0.20}& 0.23 \\ 
 \hline 
idiap & 0.25& 0.35& 0.27& 0.25& \textbf{0.24} \\ 
 \hline 
lia & 0.24& 0.34& 0.17& \textbf{0.15}& \underline{0.16} \\ 
 \hline 
uman & 0.27& 0.33& \textbf{0.18}& \underline{0.20}& 0.21 \\ 
 \hline 
unis & 0.2& 0.27& \textbf{0.07}& \underline{0.1}& \underline{0.1} \\ 
 \hline 
uoulu & \underline{0.18}& 0.23& \textbf{0.14}& \textbf{0.14} & 0.19 \\ 
 \hline  \hline 
Altogether & 0.22& 0.28& \textbf{0.17}& \underline{0.18}& 0.19 \\ 
 \hline \hline
Mobile-PC & 0.27& 0.38& \textbf{0.19}& 0.21& \underline{0.2} \\ 
 \hline 
\end{tabular} 
}
\caption{The EER values corresponding to MOBIO dataset experiments.}
\label{tab:mobio_eer}
\end{table}

The second experiment is about the cross sensor authentication, in which you enroll yourself on one device and test on another device. To see how important sensor change can be for low level features, one can look at the performance drop of the LBP features in this experiment and the previous one in Table \ref{tab:mobio_eer}. The decrease is $0.10$ for the LBP features and $0.05$ for ACA which depends on low level features, while \MACNN\ methods just have a decrease of $0.01$ in EER value. The ROC curve for this experiment is presented in Figure \ref{fig:mobio_roc_mobile}. Again, this is due to the fact that \MACNN\ has seen more variations in the large training set. Our method can also handle partial face verification if a partial face detector \cite{mahbub2016partial} is available.

\section{Mobile performance}
\begin{table*}[ht]
 \centering
 \resizebox{0.8\linewidth}{!}{
 \begin{tabular}{|c||c|c|c||c|c|c|}
 \hline
Input size & Network& Parameters& Prediction time& Network& Parameters& Prediction time \\
 \hline \hline
$128 \times 52$ & D-UpperHead& 275,360& 0.15s& W-UpperHead& 1,825,664& 0.26s \\
 \hline
$115 \times 41$ & D-BothEyes& 227,936& 0.11s& W-BothEyes& 1,447,552& 0.19s \\
 \hline
$90 \times 62$ & D-EyesNose& 244,704& 0.13s& W-EyesNose& 1,580,160& 0.22s \\
 \hline
$40 \times 56$ & D-Nose& 170,400& 0.06s& W-Nose& 988,032& 0.1s \\
 \hline
$55 \times 82$ & D-NoseMouth& 232,352& 0.10s& W-NoseMouth& 1,481,600& 0.18s \\
 \hline
$65 \times 38$ & D-Mouth& 164,448& 0.06s& W-Mouth& 939,648& 0.11s \\
 \hline
$115 \times 107$ & D-EyesNoseMouth& 441,632& 0.28s& W-EyesNoseMouth& 3,154,304& 0.48s \\
 \hline
$128 \times 45$ & D-MouthChin& 244,640& 0.13s& W-MouthChin& 1,579,904& 0.23s \\
 \hline
$62 \times 100$ & D-Ear& 256,864& 0.14s& W-Ear& 1,677,952& 0.25s \\
 \hline
$53 \times 39$ & D-Eye& 162,400& 0.06s& W-Eye& 923,264& 0.08s \\
\hline \hline
Overall & MultiDeep-\MACNN\ & 2.4M & 1.22s & MultiWide-\MACNN\ & 15.6M & 2.10s \\ 
 \hline \hline
$128 \times 128$ & BinaryDeep-Full& 584,160& 0.36s& BinaryWide-Full& 4,289,664& 0.637s \\
 \hline
\end{tabular}
}
\caption{Network size and prediction speed of the networks. The D-* means it has MultiDeep-\MACNN\ architecture and W-* means it is MultiWide-\MACNN. The Binary*-\MACNN\ network prediction times are just for one attribute. For all of them together it will be 40 times this value.}
\label{tab:performance}
\end{table*}

There is a trade-off among power consumption, authentication speed, and accuracy of the model for the task of active authentication on mobile devices. The response time is important since we do not want to freeze other running processes and create an unpleasant user experience while authenticating. Power consumption is also important because as frequent demands for charging the battery can be annoying. 

To show the effectiveness of our approach, we measure the attribute prediction speed of our networks and the battery consumption on an LG Nexus 5 device. The results are shown in Table \ref{tab:performance}. This mobile device has a quad-core QUALCOMM Snapdragon 800 clocked at 2.26 GHz and 2 GB of RAM. This specification is considered average compared to the current smartphones. We use the Tensorflow \cite{tensorflow} implementation of CNNs on Android devices.
 
We follow ACA \cite{faaa} for the performance analysis on the phone. We take one shot with the smartphone camera and feed it to the network 200 times and measure the prediction speed by looking at the average duration per frame. To measure the power usage we use PowerTutor \cite{powertutor} which registers the energy usage per running application and also in total. We do not use the camera continuously because it will bias the response time and power usage of the network. We take the image and the application works in background. The default Android processes are the only other processes that are running besides our application that runs the networks.

According to Table \ref{tab:performance} all the attributes are detected in $1.22s$ with MultiDeep-\MACNN\ running on CPU in the background without blocking other applications. MultiWide-\MACNN\ takes $2.10s$. The BinaryDeep-\MACNN\ takes $14.4s$ and BinaryWide-\MACNN\ $25.5s$. 

The MultiDeep-\MACNN\ architecture consumes $780mW$ power on average and MultiWide-\MACNN\ drains $1100mW$ of the battery power. The average battery usage of Android when it is not running the \MACNN\ networks is $600mW$ according to PowerTutor. To see how this affects the battery life, suppose the battery capacity is $C$ Watt-hours (Wh). Then 
\begin{equation}
d = \frac{C}{P_n+\beta \alpha P_d}
\end{equation}
where $d$ is the mobile device's battery life, $P_n$ is the power consumption in normal use, $P_d$ is the power usage of the attribute detection algorithm, $\beta$ is the fraction of time that the mobile device is being used, $\alpha$ is the authentication ratio constant. $\alpha$ shows how often we want to authenticate the user considering the prediction time of the algorithm, \ie we authenticate every $\frac{T_a}{\alpha}$ where $T_a$ is the prediction speed of the model. For instance, if $\alpha = 0.5$ we authenticate every $2.44s$ using MultiDeep-\MACNN\ and every $4.2s$ using MultiWide-\MACNN.

To make the feasibility of \MACNN\ clearer, suppose we authenticate the user using the MultiDeep-\MACNN\ architecture on the Nexus 5 device. We choose the MultiDeep-\MACNN\ since it performs well in the authentication task as discussed in Section \ref{sec:activeauth} and also it has a better runtime and power usage. The Nexus 5 has a $2300mAh$ battery with $3.8V$ voltage, so $C = 8.74Wh$. $P_n = 0.6W$ for the ``normal usage'' state which is when just Android 5 and the default applications are running. This gives $14.5$ hours battery life. Now if $\alpha = 1$  which means we want to authenticate with the highest speed possible and if we are using the phone all the time with $\beta=1$ then the battery life will be reduced to $6.3$ hours in the worst case. In a realistic setting with $\beta=0.2$ and $\alpha=0.5$ it becomes $12.85$ hours which is reasonable. Also, if a GPU implementation of CNNs on Android \cite{sayantan} is used, attribute prediction can happen much faster with less energy consumption. 

\label{sec:performance}
\section{Discussion and future direction}
\label{sec:discussion}
We proposed a feasible multi-task DCNN architecture to extract accurate and describable facial attributes on mobile devices. Each network predicts multi facial attributes from a given face component by mapping it to a shared embedding space. We showed that our attribute prediction performance is comparable to state-of-the art. We explored the embedding space and illustrated that we can extract new attributes by looking at subspace clusters of this space. We also showed that our networks perform attribute-based authentication better than the previously proposed method \cite{mybtas,faaa}. Finally, we analyzed the feasibility of our method by performing battery usage and prediction speed experiments on an actual mobile device. 

In the future, we plan to jointly train this ensemble of networks for the task of face verification and attribute prediction to get a more discriminative embedding space to gain better authentication performance.
\section*{Acknowledgement}
	This work was supported by cooperative agreement FA8750-13-2-0279 from DARPA.
	
{\small
\bibliographystyle{ieee}
\bibliography{submission_example}

\begin{thebibliography}{10}\itemsep=-1pt

\bibitem{tensorflow}
M.~Abadi, A.~Agarwal, P.~Barham, E.~Brevdo, Z.~Chen, C.~Citro, G.~S. Corrado,
  A.~Davis, J.~Dean, M.~Devin, S.~Ghemawat, I.~Goodfellow, A.~Harp, G.~Irving,
  M.~Isard, Y.~Jia, R.~Jozefowicz, L.~Kaiser, M.~Kudlur, J.~Levenberg,
  D.~Man\'{e}, R.~Monga, S.~Moore, D.~Murray, C.~Olah, M.~Schuster, J.~Shlens,
  B.~Steiner, I.~Sutskever, K.~Talwar, P.~Tucker, V.~Vanhoucke, V.~Vasudevan,
  F.~Vi\'{e}gas, O.~Vinyals, P.~Warden, M.~Wattenberg, M.~Wicke, Y.~Yu, and
  X.~Zheng.
\newblock {TensorFlow}: Large-scale machine learning on heterogeneous systems,
  2015.
\newblock Software available from tensorflow.org.

\bibitem{lbp}
T.~Ahonen, A.~Hadid, and M.~Pietikainen.
\newblock Face description with local binary patterns: Application to face
  recognition.
\newblock {\em Pattern Analysis and Machine Intelligence, IEEE Transactions
  on}, 28(12):2037--2041, 2006.

\bibitem{antal2016biometric}
M.~Antal and L.~Z. Szab{\'o}.
\newblock Biometric authentication based on touchscreen swipe patterns.
\newblock {\em Procedia Technology}, 22:862--869, 2016.

\bibitem{asthana}
A.~Asthana, S.~Zafeiriou, S.~Cheng, and M.~Pantic.
\newblock Robust discriminative response map fitting with constrained local
  models.
\newblock In {\em Computer Vision and Pattern Recognition (CVPR), 2013 IEEE
  Conference on}, pages 3444--3451. IEEE, 2013.

\bibitem{libsvm}
C.-C. Chang and C.-J. Lin.
\newblock Libsvm: a library for support vector machines.
\newblock {\em ACM Transactions on Intelligent Systems and Technology (TIST)},
  2(3):27, 2011.

\bibitem{clarke2007authenticating}
N.~L. Clarke and S.~M. Furnell.
\newblock Authenticating mobile phone users using keystroke analysis.
\newblock {\em International Journal of Information Security}, 6(1):1--14,
  2007.

\bibitem{Jain_AA_ICB2015}
D.~Crouse, H.~Han, D.~Chandra, B.~Barbello, and A.~K. Jain.
\newblock Continuous authentication of mobile user: Fusion of face image and
  inertial measurement unit data.
\newblock In {\em International Conference on Biometrics}, 2015.

\bibitem{MobileGait}
M.~Derawi, C.~Nickel, P.~Bours, and C.~Busch.
\newblock Unobtrusive user-authentication on mobile phones using biometric gait
  recognition.
\newblock In {\em International Conference on Intelligent Information Hiding
  and Multimedia Signal Processing}, pages 306--311, Oct 2010.

\bibitem{ssc}
E.~Elhamifar and R.~Vidal.
\newblock Sparse subspace clustering.
\newblock In {\em Computer Vision and Pattern Recognition, 2009. CVPR 2009.
  IEEE Conference on}, pages 2790--2797. IEEE, 2009.

\bibitem{UMDAA}
M.~E. Fathy, V.~M. Patel, and R.~Chellappa.
\newblock Face-based active authentication on mobile devices.
\newblock In {\em IEEE International Conference on Acoustics, Speech and Signal
  Processing}, 2015.

\bibitem{Continuous_HST2012}
T.~Feng, Z.~Liu, K.-A. Kwon, W.~Shi, B.~Carbunar, Y.~Jiang, and N.~Nguyen.
\newblock Continuous mobile authentication using touchscreen gestures.
\newblock In {\em IEEE Conference on Technologies for Homeland Security}, pages
  451--456, Nov 2012.

\bibitem{Touchalytics}
M.~Frank, R.~Biedert, E.~Ma, I.~Martinovic, and D.~Song.
\newblock Touchalytics: On the applicability of touchscreen input as a
  behavioral biometric for continuous authentication.
\newblock {\em IEEE Transactions on Information Forensics and Security},
  8(1):136--148, Jan 2013.

\bibitem{Lex_stylometry}
L.~Fridman, S.~Weber, R.~Greenstadt, and M.~Kam.
\newblock Active authentication on mobile devices via stylometry, gps location,
  web browsing behavior, and application usage patterns.
\newblock {\em IEEE Systems Journal}, 2015.

\bibitem{face_eye_mobile}
A.~Hadid, J.~Heikkila, O.~Silven, and M.~Pietikainen.
\newblock Face and eye detection for person authentication in mobile phones.
\newblock In {\em ACM/IEEE International Conference on Distributed Smart
  Cameras}, pages 101--108, Sept 2007.

\bibitem{nq}
N.~M. Inc.
\newblock Nearly one in three consumers who have lost their mobile devices
  still do not lock them, new survey shows.
\newblock
  "http://www.prnewswire.com/news-releases/nearly-one-in-three-consumers-who-have-lost-their-mobile-devices-still-do-not-lock-them-new-survey-shows-200410151.html",
  2013.

\bibitem{softbiometric}
A.~K. Jain, S.~C. Dass, and K.~Nandakumar.
\newblock Can soft biometric traits assist user recognition?
\newblock In {\em Defense and Security}, pages 561--572. International Society
  for Optics and Photonics, 2004.

\bibitem{adam}
D.~Kingma and J.~Ba.
\newblock Adam: A method for stochastic optimization.
\newblock {\em arXiv preprint arXiv:1412.6980}, 2014.

\bibitem{alexnet}
A.~Krizhevsky, I.~Sutskever, and G.~E. Hinton.
\newblock Imagenet classification with deep convolutional neural networks.
\newblock In {\em Advances in neural information processing systems}, pages
  1097--1105, 2012.

\bibitem{facetracer}
N.~Kumar, P.~N. Belhumeur, and S.~K. Nayar.
\newblock {F}ace{T}racer: {A} {S}earch {E}ngine for {L}arge {C}ollections of
  {I}mages with {F}aces.
\newblock In {\em European Conference on Computer Vision (ECCV)}, pages
  340--353, Oct 2008.

\bibitem{hassner}
G.~Levi and T.~Hassner.
\newblock Age and gender classification using convolutional neural networks.
\newblock In {\em IEEE Conf. on Computer Vision and Pattern Recognition (CVPR)
  workshops}, June 2015.

\bibitem{celeba}
Z.~Liu, P.~Luo, X.~Wang, and X.~Tang.
\newblock Deep learning face attributes in the wild.
\newblock In {\em Proceedings of International Conference on Computer Vision
  (ICCV)}, December 2015.

\bibitem{mahbub2016partial}
U.~Mahbub, V.~M. Patel, D.~Chandra, B.~Barbello, and R.~Chellappa.
\newblock Partial face detection for continuous authentication.
\newblock {\em arXiv preprint arXiv:1603.09364}, 2016.

\bibitem{mobio}
C.~McCool, S.~Marcel, A.~Hadid, M.~Pietikainen, P.~Matejka, J.~Cernocky,
  N.~Poh, J.~Kittler, A.~Larcher, C.~Levy, D.~Matrouf, J.-F. Bonastre,
  P.~Tresadern, and T.~Cootes.
\newblock Bi-modal person recognition on a mobile phone: using mobile phone
  data.
\newblock In {\em IEEE ICME Workshop on Hot Topics in Mobile Multimedia}, July
  2012.

\bibitem{niinuma2010soft}
K.~Niinuma, U.~Park, and A.~K. Jain.
\newblock Soft biometric traits for continuous user authentication.
\newblock {\em Information Forensics and Security, IEEE Transactions on},
  5(4):771--780, 2010.

\bibitem{vgg-face}
O.~M. Parkhi, A.~Vedaldi, and A.~Zisserman.
\newblock Deep face recognition.
\newblock In {\em British Machine Vision Conference}, 2015.

\bibitem{early_stopping}
L.~Prechelt.
\newblock Automatic early stopping using cross validation: quantifying the
  criteria.
\newblock {\em Neural Networks}, 11(4):761--767, 1998.

\bibitem{context_AA}
A.~Primo, V.~Phoha, R.~Kumar, and A.~Serwadda.
\newblock Context-aware active authentication using smartphone accelerometer
  measurements.
\newblock In {\em Computer Vision and Pattern Recognition Workshops (CVPRW),
  2014 IEEE Conference on}, pages 98--105, June 2014.

\bibitem{mybtas}
P.~Samangouei, V.~M. Patel, and R.~Chellappa.
\newblock Attribute-based continuous user authentication on mobile devices.
\newblock In {\em Biometrics Theory, Applications and Systems (BTAS), 2015 IEEE
  7th International Conference on}, pages 1--8, Sept 2015.

\bibitem{faaa}
P.~Samangouei, V.~M. Patel, and R.~Chellappa.
\newblock Facial attributes for active authentication on mobile devices.
\newblock {\em Image and Vision Computing}, 2016.

\bibitem{sayantan}
S.~Sarkar, V.~M. Patel, and R.~Chellappa.
\newblock Deep feature-based face detection on mobile devices.
\newblock {\em arXiv preprint arXiv:1602.04868}, 2016.

\bibitem{facenet}
F.~Schroff, D.~Kalenichenko, and J.~Philbin.
\newblock Facenet: A unified embedding for face recognition and clustering.
\newblock In {\em Proceedings of the IEEE Conference on Computer Vision and
  Pattern Recognition}, pages 815--823, 2015.

\bibitem{deepid}
Y.~Sun, Y.~Chen, X.~Wang, and X.~Tang.
\newblock Deep learning face representation by joint
  identification-verification.
\newblock In {\em Advances in Neural Information Processing Systems}, pages
  1988--1996, 2014.

\bibitem{sun2014deep}
Y.~Sun, Y.~Chen, X.~Wang, and X.~Tang.
\newblock Deep learning face representation by joint
  identification-verification.
\newblock In {\em Advances in Neural Information Processing Systems}, pages
  1988--1996, 2014.

\bibitem{deepface}
Y.~Taigman, M.~Yang, M.~Ranzato, and L.~Wolf.
\newblock Deepface: Closing the gap to human-level performance in face
  verification.
\newblock In {\em Proceedings of the IEEE Conference on Computer Vision and
  Pattern Recognition}, pages 1701--1708, 2014.

\bibitem{omp}
J.~A. Tropp and A.~C. Gilbert.
\newblock Signal recovery from random measurements via orthogonal matching
  pursuit.
\newblock {\em Information Theory, IEEE Transactions on}, 53(12):4655--4666,
  2007.

\bibitem{src}
J.~Wright, A.~Y. Yang, A.~Ganesh, S.~S. Sastry, and Y.~Ma.
\newblock Robust face recognition via sparse representation.
\newblock {\em Pattern Analysis and Machine Intelligence, IEEE Transactions
  on}, 31(2):210--227, 2009.

\bibitem{Heng_FG2015_Fusion}
H.~Zhang, V.~M. Patel, and R.~Chellappa.
\newblock Robust multimodal recognition via multitask multivariate low-rank
  representations.
\newblock In {\em IEEE International Conference on Automatic Face and Gesture
  Recognition}. IEEE, 2015.

\bibitem{Heng_WACV2015}
H.~Zhang, V.~M. Patel, M.~E. Fathy, and R.~Chellappa.
\newblock Touch gesture-based active user authentication using dictionaries.
\newblock In {\em IEEE Winter conference on Applications of Computer Vision}.
  IEEE, 2015.

\bibitem{powertutor}
L.~Zhang, B.~Tiwana, Z.~Qian, Z.~Wang, R.~P. Dick, Z.~M. Mao, and L.~Yang.
\newblock Accurate online power estimation and automatic battery behavior based
  power model generation for smartphones.
\newblock In {\em Proceedings of the eighth IEEE/ACM/IFIP international
  conference on Hardware/software codesign and system synthesis}, pages
  105--114. ACM, 2010.

\bibitem{zhang2014panda}
N.~Zhang, M.~Paluri, M.~Ranzato, T.~Darrell, and L.~Bourdev.
\newblock Panda: Pose aligned networks for deep attribute modeling.
\newblock In {\em Computer Vision and Pattern Recognition (CVPR), 2014 IEEE
  Conference on}, pages 1637--1644. IEEE, 2014.

\bibitem{zhong2014sensor}
Y.~Zhong and Y.~Deng.
\newblock Sensor orientation invariant mobile gait biometrics.
\newblock In {\em Biometrics (IJCB), 2014 IEEE International Joint Conference
  on}, pages 1--8. IEEE, 2014.

\end{thebibliography}
}

\end{document}